\documentclass[10pt,twocolumn,letterpaper]{article}

\usepackage{cvpr}
\usepackage{times}
\usepackage{epsfig}
\usepackage{graphicx}
\usepackage{amsmath}
\usepackage{amssymb}
\usepackage{algpseudocode}
\usepackage{soul}
\usepackage{subcaption,graphicx}

\usepackage[pagebackref=true,breaklinks=true,letterpaper=true,colorlinks,bookmarks=false]{hyperref}

\cvprfinalcopy 

\def\cvprPaperID{} 

\ifcvprfinal\fi
\begin{document}

\title{Fast and Furious: Real Time End-to-End 3D Detection, Tracking and Motion Forecasting with a Single Convolutional Net}

\author{Wenjie Luo, Bin Yang and Raquel Urtasun\\
Uber ATG, University of Toronto\\
{\tt\small \{wenjie, byang10, urtasun\}@uber.com}
}

\maketitle

\begin{abstract}

In this paper we propose a novel  deep neural network that is able to jointly reason about 3D detection, tracking and motion forecasting  given data captured by a 3D sensor.
By jointly reasoning about these tasks, our  holistic approach is  more robust to occlusion as well as sparse data at range.
Our approach performs 3D convolutions across space and time over a bird's eye view representation of the 3D world, which  is very efficient in terms of both  memory and computation.
Our experiments on a new very large scale dataset captured  in several north american cities,   show that we can outperform the state-of-the-art by a large margin.
Importantly, by sharing computation we can perform all  tasks in as little as 30 ms.

\end{abstract}

\section{Introduction}

Modern approaches to self-driving  divide the problem into four steps:  detection, object  tracking,  motion forecasting and motion planning. A cascade approach is typically used where the output of the detector is used as input to the tracker, and its output is fed to a motion forecasting algorithm that estimates where traffic participants are going to move in the next few seconds.  This is in turn fed to the motion planner that estimates the final trajectory of the ego-car.
These modules are usually learned independently, and uncertainty is usually rarely propagated.
This can result in catastrophic failures as downstream processes cannot recover from errors that appear at the beginning of the pipeline.

\begin{figure}[t]
\begin{center}
   \includegraphics[width=1\linewidth]{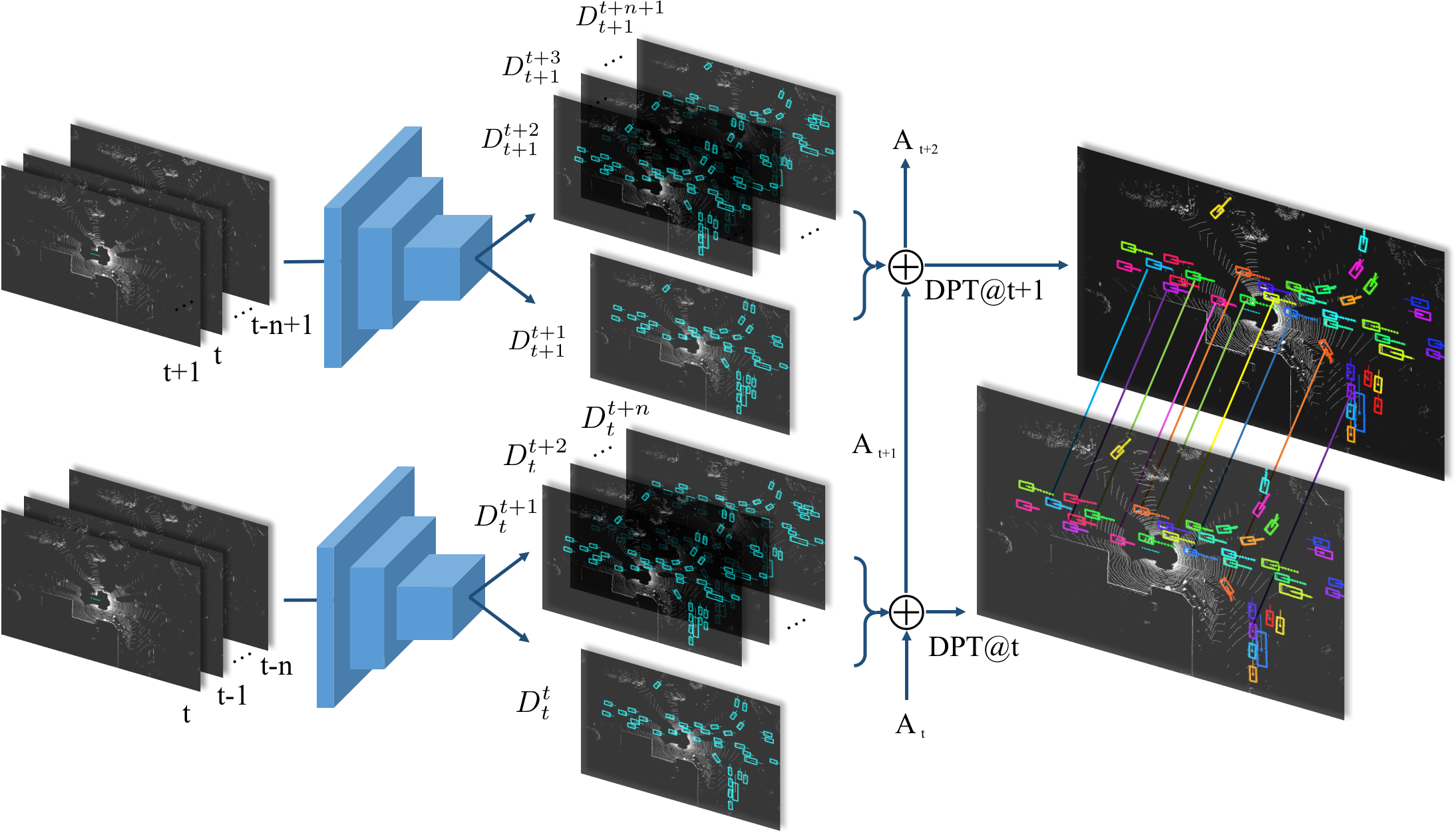}
\end{center}
   \caption{\textbf{Overview of our approach: }  Our $FaF$ network takes multiple frames as input and  performs detection, tracking and  motion forecasting. }
\label{fig:overview}
\end{figure}

In contrast, in this paper we propose an end-to-end fully convolutional approach that performs simultaneous 3D detection, tracking and motion forecasting by exploiting spatio-temporal information captured by a  3D sensor.
We argue that this is important as tracking and prediction can help object detection.
For example, leveraging tracking and prediction information can  reduce detection false negatives when dealing with occluded or far away objects.  False positives can also be reduced by accumulating evidence over time.
Furthermore, our approach is very efficient as it shares computations between all these tasks.  This is extremely  important for autonomous driving where latency can be fatal.

We take advantage of 3D sensors and design a network that operates on a  bird-eye-view (BEV) of the 3D world. This representation respects the 3D nature of the sensor data, making the learning process easier as the network can exploit priors about the typical sizes of  objects.

Our approach is   a   one-stage detector that takes a 4D tensor created from multiple consecutive temporal frames as input and  performs  3D convolutions  over space and time to extract accurate 3D bounding boxes. 
Our model produces bounding boxes not only at the current frame but also multiple timestamps into the future. We  decode the tracklets from these predictions by a simple pooling operation that combine  the evidence from past and current predictions.

We demonstrate the effectiveness of our model   on a very large scale dataset captured from multiple vehicles driving in North-america and show that our approach significantly outperforms the state-of-the-art. Furthermore, all tasks take as little as 30ms. 

\section{Related Work}

\paragraph{2D Object Detection:} Over the last few years many methods that exploit convolutional neural networks to produce accurate 2D object detections, typically from a single image, have been developed.
These approaches typically fell into two categories depending on whether they exploit a first step dedicated to create object proposals.
Modern {\it two-stage detectors} \cite{ren2015faster, he2017maskrcnn, dai2016r, huang2016speed}, utilize region proposal networks (RPN)  to learn the region of interest (RoI) where potential objects are located. In a second stage the final  bounding box locations are predicted from  features that are average-pooled over the proposal RoI.
{Mask-RCNN \cite{he2017maskrcnn} also took this approach, but  used  RoI aligned features addressing the{boundary and quantization effect of RoI pooling.
Furthermore, they added an additional segmentation branch to take advantage of dense pixel-wise  supervision, achieving state-of-the-art results on both 2D image detection and instance segmentation.  On the other hand  {\it one-stage detectors} skip the  proposal generation step, and instead  learn a  network that directly produces   object bounding boxes. Notable examples are  YOLO \cite{redmon2016yolo9000}, SSD \cite{liu2016ssd} and RetinaNet \cite{lin2017focal}}. One-stage detectors  are computationally very appealing  and are typically real-time, especially with the help of recently proposed architectures, \eg MobineNet \cite{howard2017mobilenets}, SqueezeNet \cite{wu2016squeezedet}.  One-stage detectors were outperformed significantly by two stage-approaches until Lin \etal \cite{lin2017focal} shown state-of-the-art results by exploiting a focal loss and dense predictions.

\paragraph{3D Object Detection:}
In robotics applications such as autonomous driving we are interested in detecting objects in 3D space.
The ideas behind modern  2D image  detectors can  be transferred to 3D object detection. Chen \etal \cite{chen20173d} used stereo images to perform 3D detection. Li \cite{li20163d} used 3D point cloud data and proposed to use 3D convolutions on a voxelized representation of point clouds. Chen \etal \cite{chen2016multi} combined image and 3D point clouds with a fusion network. They exploited 2D convolutions in BEV, however, they used hand-crafted height features as input. They achieved promising results on KITTI \cite{Geiger2012CVPR} but only ran at 360ms per frame due to heavy feature computation on both 3D point clouds and images. This is very slow, particularly if we are interested in extending these techniques to handle temporal data. 

\paragraph{Object Tracking:} Over the past few decades many approaches have been develop for object tracking. In this section we briefly review the use of deep learning in tracking.
Pretrained CNNs were used  to extract features and perform tracking with correlation \cite{ma2015hierarchical} or regression \cite{wang2015visual, held2016learning}.
Wang and Yeung \cite{wang2013learning} used an autoencoder to learn a good feature representation that helps tracking. Tao \etal \cite{tao2016siamese} used siamese matching networks to perform tracking.
Nam and Han \cite{nam2016learning}  finetuned a CNN  at inference time to track object within the same video.

\paragraph{Motion Forecasting:} This is the problem of predicting  where each object will be in the future given multiple past frames. Lee \etal \cite{lee2017desire} proposed to use recurrent networks  for long term prediction. Alahi \etal \cite{alahi2016social} used LSTMs to model the interaction between pedestrian and perform prediction accordingly. Ma \etal \cite{ma2017forecasting} proposed to utilize concepts from game theory to model the interaction between pedestrian while  predicting future trajectories. Some work has also focussed on short term prediction of dynamic objects \cite{gong2011multi, pellegrini2009you}. \cite{walker2016uncertain} performed prediction for dense pixel-wise short-term trajectories using variational autoencoders. \cite{srivastava2015unsupervised, mathieu2015deep} focused on predicting the next future frames given a video, without explicitly reasoning about  per pixel motion.

\paragraph{Multi-task Approaches:}
Feichtenhofer \etal \cite{feichtenhofer2017detect} proposed to do detection and tracking jointly from video. They model the displacement of corresponding objects between two input images during training  and decode them into object tubes during inference time.

Different from all the above work, in this paper we propose a single network that takes advantage of temporal information and tackles the problem of 3D detection, short term motion forecasting and tracking  in the scenario of autonomous driving.
While temporal information provides us with important  cues for motion forecasting, holistic reasoning   allows us to better propagate the uncertainty throughout the network, improving our estimates.
Importantly, our model is super efficient and runs real-time at 33 FPS.

\section{Joint 3D Detection, Tracking and Motion Forecasting}

In this work, we focus on detecting objects by exploiting a  sensor which produces  3D point clouds.
Towards this goal, we develop a one-stage detector which takes as input multiple frames and produces detections, tracking and short term motion forecasting  of the objects' trajectories into the future.
Our input representation is a 4D tensor encoding  an occupancy grid of the 3D space over several  time frames.
We  exploit 3D convolutions over space and time to produce fast and accurate predictions.
As  point cloud data is inherently sparse in 3D space, our approach saves lots of computation as compared to doing 4D convolutions over 3D space and time. 
We name our approach Fast and Furious (FaF), as it is able to create very accurate estimates in as little as 30 ms.

In the following, we  first describe  our  data parameterization in Sec.~\ref{sec:data} including voxelization and how we incorporate temporal information. In Sec.~\ref{sec:model}, we present our model's architecture, follow by  the objective we use for training the network  (Sec.~\ref{sec:loss}).

\subsection{Data Parameterization} \label{sec:data}

\begin{figure}[t]
\begin{center}
   \includegraphics[width=0.8\linewidth]{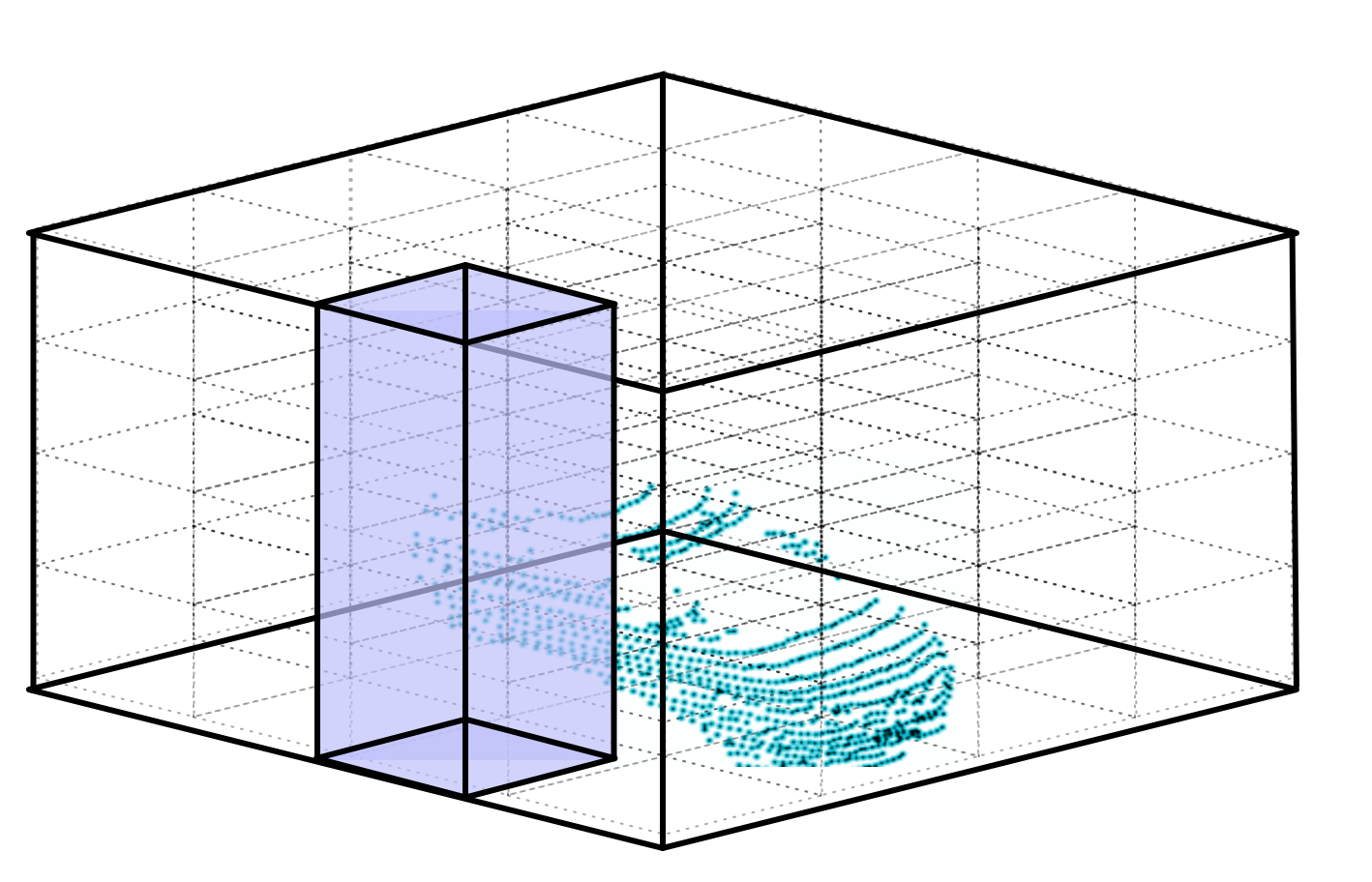}
\end{center}
   \caption{\textbf{Voxel Representation:} using height directly as input feature.}
\label{fig:voxel}
\end{figure}

\begin{figure}[t]
\begin{center}
   \includegraphics[width=0.9\linewidth]{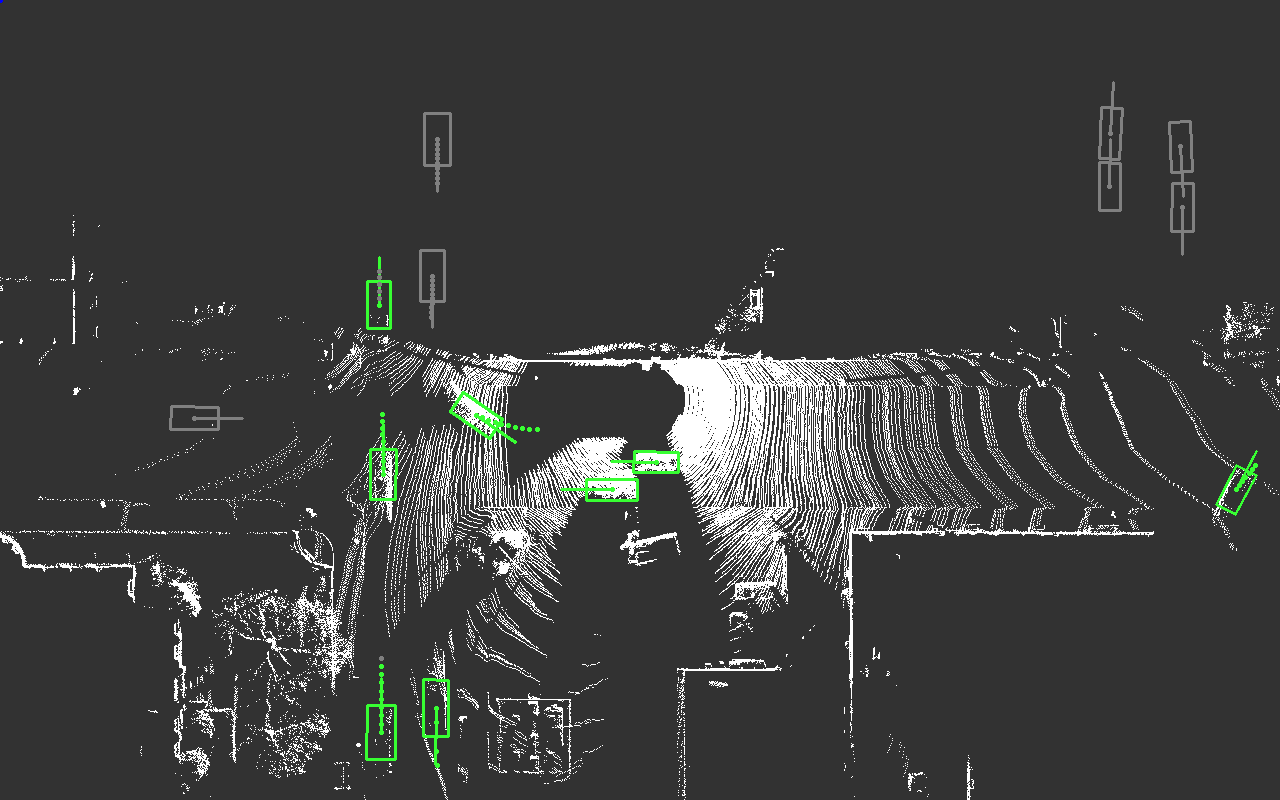}
\end{center}
   \caption{\textbf{Overlay temporal \& motion forecasting data. Green: bbox w/ 3D point. Grey: bbox w/o 3D point}}
\label{fig:tensor}
\end{figure}


\begin{figure*}[htp]
  \centering
  \begin{subfigure}{0.49\textwidth}
    \centering
    \includegraphics[width=0.9\linewidth]{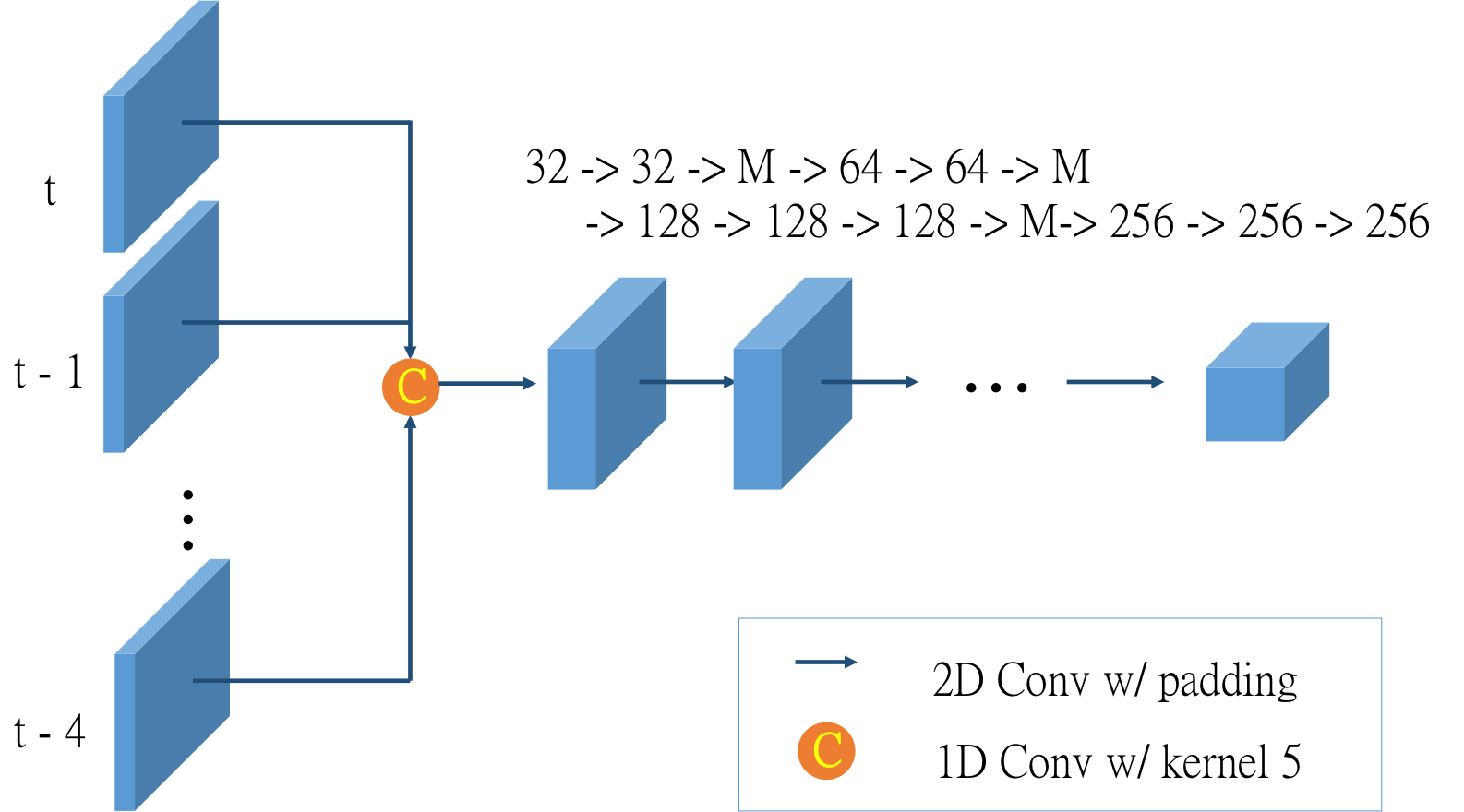}
    \caption{Early fusion}
  \end{subfigure}
  \begin{subfigure}{0.49\textwidth}
    \centering
    \includegraphics[width=0.9\linewidth]{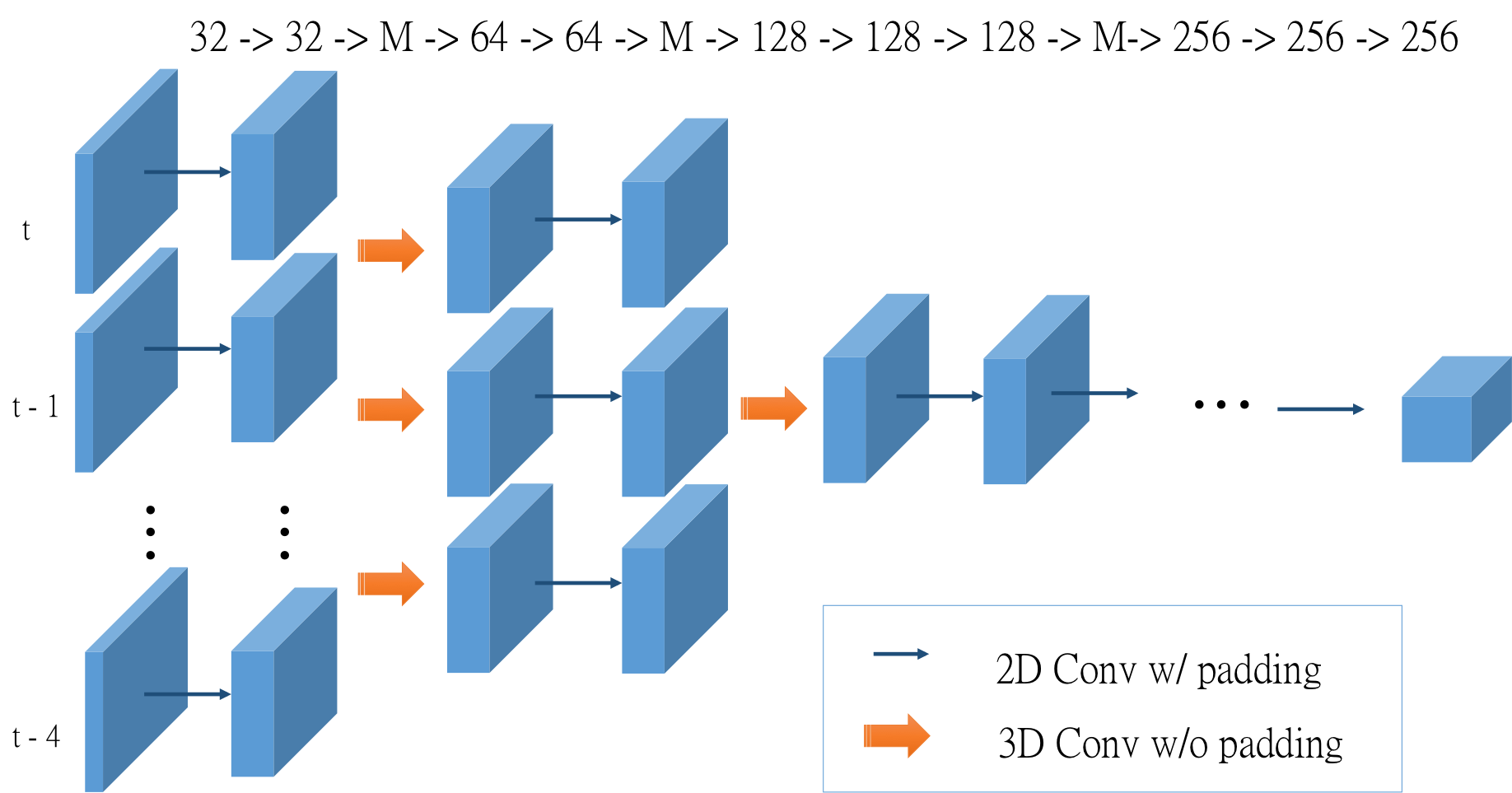}
    \caption{Later fusion}
  \end{subfigure}
\caption {\textbf{Modeling temporal information}}
\label{fig:model}
\end{figure*}

In this section, we first describe our single frame representation of the world. We then extend our representation to exploit multiple frames.

\paragraph{Voxel Representation:} In contrast to image detection where the input is a dense RGB image, point cloud data is inherently sparse and provides geometric  information about the 3D  scene.
In order to get a representation where convolutions can be easily applied, we quantize the 3D world to form a 3D voxel  grid.  We then assign a binary indicator for each voxel encoding whether the voxel is occupied. We say a voxel is  occupied   if there exists at least one  point in the voxel's 3D space.
As the grid is a regular lattice, convolutions can be directly used. We do not utilize 3D convolutions on our single frame representation as this operation will waste most computation since the grid is very sparse, i.e., most of the voxels are not occupied.
Instead, we performed 2D convolutions and treat the height dimension as the channel dimension. This allows the network to learn to extract information in the height dimension.
This contrast approaches such as MV3D \cite{chen2016multi}, which perform  quantization on the x-y plane and  generate a representation of the z-dimension by computing hand-crafted height statistics.
Note that if our  grid's resolution  is high, our approach  is equivalent to applying convolution on every single point without loosing any information. We refer the reader to
Fig.~\ref{fig:voxel} for an illustration of how we construct the 3D tensor from 3D point cloud data.

\paragraph{Adding Temporal Information:} In order to perform motion forecasting, it is crucial to consider temporal information. Towards this goal,  we take all the 3D points from the past $n$ frames and perform a change of coordinates to represent then   in the current vehicle coordinate system. This is important in order to undo the ego-motion of the vehicle where the sensor is mounted.

After performing this transformation, we compute the voxel representation for each frame.
Now that   each frame is represented as a 3D  tensor, we can  append multiple frames'  along a new temporal dimension to create a 4D tensor.
This not only provides more 3D points as a whole, but also gives cues about vehicle's heading and  velocity  enabling us to do motion forecasting.
As shown in Fig.~\ref{fig:tensor}, where  for visualization purposes we overlay multiple frames,  static objects are well aligned while dynamic objects have `shadows'  which represents their motion.

\subsection{Model Formulation} \label{sec:model}

Our single-stage detector takes a 4D input tensor and regresses directly to object bounding boxes at different timestamps without using region proposals.
We investigate two different ways to exploit the temporal dimension on our 4D  tensor: early fusion and late fusion.
They represent a tradeoff between accuracy and efficiency, and they differ on at which level the temporal dimension is aggregated.

\paragraph{Early Fusion:}
Our first approach  aggregates temporal information at the very first layer. As a consequence  it runs as  fast as using the single frame detector. However, it might lack the ability to capture complex temporal features as this is equivalent to producing a single point cloud from all frames, but weighting the contribution of the different timestamps differently.
In particular, as shown in Fig.~\ref{fig:model}, given a 4D input tensor, we first use a 1D convolution with kernel size $n$ on temporal dimension to reduce the temporal dimension from $n$ to 1. We share the   weights  among all feature maps, \ie, also known as group convolution.
We then perform convolution and max-pooling following VGG16 \cite{simonyan2014very} with each layer number of feature maps reduced by half. Note that we remove the last convolution group in VGG16, resulting in only 10 convolution layers.

\paragraph{Late Fusion:}
In this case, we gradually merge the temporal information. This allows the model to capture high level motion features.
We use the same number of convolution layers and feature maps  as in the early fusion model, but instead perform 3D convolution with kernel size $3\times 3 \times 3$ for 2 layers without padding on temporal dimension, which reduces the temporal dimension from $n$ to 1, and then perform 2D spatial convolution with kernel size $3\times 3$ for other layers.
We refer the reader to Fig.~\ref{fig:model} for an illustration of our architecture.

We then add two branches of convolution layers as shown in Fig.~\ref{fig:prediction}.
The first one performs binary classification to predict the probability of being a vehicle.
The second one predicts the bounding box over the current frame as well as $n-1$ frames into the future.
Motion forecasting is possible as our approach exploits multiple frames as input, and thus can learn to estimate useful features such as velocity and acceleration.

Following  SSD \cite{liu2016ssd}, we  use multiple predefined boxes for each feature map location.
As we utilize a BEV representation which is metric, our network can exploit  priors about physical sizes of  objects.   Here we use boxes corresponding to 5 meters in the real world with aspect ratio of $1:1, 1:2, 2:1, 1:6, 6:1$ and 8 meters with aspect ratio of $1:1$.
In total there are 6 predefined boxes per feature map location denoted as $a_{i,j}^k$ where $i = 1, ..., I, j = 1, ..., J$ is the location in the feature map  and $k = 1, ..., K$ ranges over the  predefined boxes (i.e., size and aspect ratio).
Using multiple predefined boxes allows us to reduce the variance of regression target, thus makes the network easy to train. Notice that we do not use predefined heading angles. Furthermore we use  both $\sin$ and $\cos$ values to avoid the $180$ degrees ambiguity.

In particular, for each predefined box $a_{i,j}^k$, our network predicts the corresponding normalized location offset $\hat{l_x}, \hat{l_y}$, log-normalized sizes $\hat{s_w}, \hat{s_h}$ and heading parameters $\hat{a}_{\sin}, \hat{a}_{\cos}$.

\begin{figure}[t]
\begin{center}
   \includegraphics[width=\linewidth]{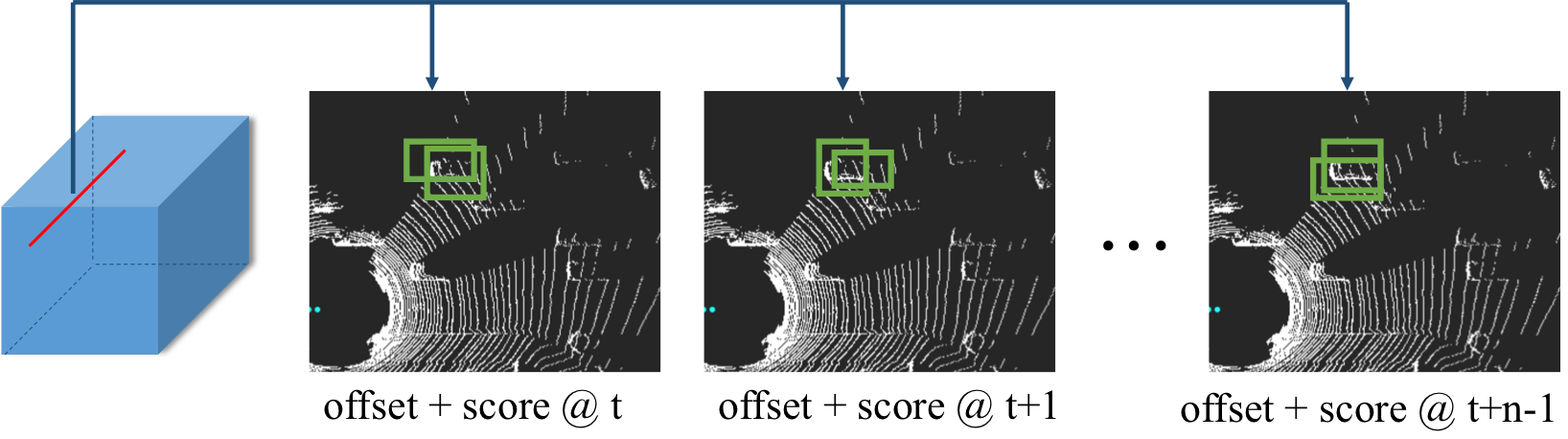}
\end{center}
   \caption{\textbf{Motion forecasting}}
\label{fig:prediction}
\end{figure}

\paragraph{Decoding Tracklets:}  At each timestamp, our model outputs the detection bounding boxes for $n$ timestamps. Reversely, each timestamp will have current detections as well as $n-1$ past predictions. Thus we can aggregate the information for the past  to produce  accurate tracklets without solving any trajectory based optimization problem. Note that  if detection and motion forecasting are perfect, we can decode perfect tracklets.
In practice, we use average as aggregation function. When there is overlap between detections from current and past's future predictions, they are considered to be the same object and their bounding boxes will simply be averaged.
Intuitively, the aggregation process helps particularly when  we have strong past predictions but no current evidence, e.g., if the object  is currently occluded or a false negative from detection.
This allow us to track through occlusions over multiple frames.
On the other hand, when we have strong current evidence but not prediction from the past, then there is evidence for a new object.

\subsection{Loss Function and Training}\label{sec:loss}
We train the network to minimize a combination of  classification and regression loss. In the case of regression we include  both the current frame as well as  our $n$ frames forecasting  into the future. That is 
\begin{equation}
\label{eq:loss}
\ell (w) = \sum \left(\alpha \cdot \ell_\text{cla}(w) + \sum\limits_{i=t,t+1,...,t+n} \ell_\text{reg}^{t}(w) \right)
\end{equation}
where $t$ is the current frame and $w$ represents the model parameters.

We employ as classification loss binary cross-entropy computed over all locations and predefined boxes:
\begin{equation}
  \ell_\text{cla}(w)  = \sum_{i,j,k} q_{i,j,k}\log p_{i,j,k}(w)
\end{equation}
Here $i,j,k$ are the indices on feature map locations and predefined box identity, $q_{i,j,k}$ is the class label (\ie $q_{i,j,k}$ =1 for vehicle and 0 for background) and $p_{i,j,k}$ is the predicted probability for vehicle.

In order to define the regression loss for our detections and future predictions,  we first need to find their associated ground truth.
We defined their  correspondence by matching each predefined box against all ground truth boxes. In particular, for each predicted box,  we first find the ground truth box with biggest overlap in terms of intersection over union (IoU). If the IoU is bigger than a fixed  threshold ($0.4$ in practice), we assign this ground truth box as $\bar{a}_{i,j}^k$ and assign 1 to its corresponding label $q_{i,j,k}$. Following SSD \cite{liu2016ssd}, if there exist a ground truth box not assigned to any predefined box, we will assign it to its highest overlapping predefined box ignoring the fixed threshold. Note that multiple predefined boxes can be associated to the same ground truth, and some predefined boxes might not have any correspondent ground truth box, meaning their $q_{i,j,k} = 0$.

Thus we  define the regression targets as 
\begin{align*}
  l_x &= \frac{x - x^{GT}}{w^{GT}} & l_y &= \frac{y - y^{GT}}{h^{GT}} \\
  s_w &= \log \frac{w}{w^{GT}} & s_h &= \log \frac{h}{h^{GT}} \\
  a_{sin} &= \sin(\theta^{GT}) & a_{cos} &= \cos(\theta^{GT})
\end{align*}
We use a weighted smooth L1 loss over all regression targets
where smooth L1 is defined as:
\begin{equation}
  \textrm{smooth}_{L_1}(\hat{x}, x) =
  \begin{cases}
    \frac{1}{2} (\hat{x} - x)^2& \text{if } |\hat{x} - x| < 1\\
    |\hat{x} - x| - \frac{1}{2}& \text{otherwise}
  \end{cases}
\end{equation}

\paragraph{Hard Data Mining} Due to the imbalance of positive and negative samples, we use hard negative mining during training. We define positive samples as those predefined boxes having corresponding ground truth box, \ie,  $q_{i,j,k} = 1$.
For negative samples, we rank all  candidates by their predicted score $p_{i,j,k}$ from the classification branch and take the top  negative samples with a ration of  3 in practice. 

\begin{table*}[t]
\setlength{\tabcolsep}{1mm}
\centering
\begin{tabular}{| c || c | c | c | c | c || c|}
		\hline
        IoU                                     & 0.5   & 0.6   & 0.7  & 0.8   & 0.9  & Time [ms] \\ \hline\hline
				SqueezeNet\_v1.1 \cite{iandola2016squeezenet} & 85.80 & 81.06 & 69.97 & 43.20 & 3.70 & 9 \\ \hline
				SSD \cite{liu2016ssd} 								  & 90.23 & 86.76 & 77.92 & 52.39 & 5.87 & 23 \\ \hline
				MobileNet \cite{howard2017mobilenets}   & 90.56 & 87.05 & 78.39 & 52.10 & 5.64 & 65 \\ \hline\hline
				FaF                                     & \bf93.24 & \bf90.54 & \bf83.10 & \bf61.61 & \bf11.83 & 30 \\ \hline
	\end{tabular}
	\caption{Detection performance on $144\times 80$ meters region, with object having $\ge$ 3 number 3D points}
	\label{tab:pr}
\end{table*}

\section{Experimental Evaluation}

Unfortunately there is no publicly available dataset which evaluates 3D detection, tracking and motion forecasting.
We thus collected a very large scale dataset in order to benchmark our approach.
It is 2 orders of magnitude bigger than datasets such as KITTI \cite{Geiger2012CVPR}.

\paragraph{Dataset:} Our dataset is collected by a roof-mounted LiDAR on top of a vehicle driving around several North-American cities. It  consists of 546,658 frames collected from 2762 different scenes. Each scene consists of  a continuous sequence. Our validation set consists of 5,000 samples collected from 100 scenes, i.e., 50 continuous frames are taken from each sequence. There is no overlap between the geographic area  where the  training and  validation are collected, in order to showcase strong generalization.
Our labels might contain vehicles with no 3D point on them as the labelers have access to the full sequence in order to provide accurate annotations. Our labels contain 3D rotated bounding box as well as track id for each vehicle.

\paragraph{Training Setup:} At training time, we use a spatial X-Y region of size $144\times 80$ meters, where each grid cell is  $0.2\times 0.2$ meters. On the height dimension, we take the range from -2 to 3.5 meters with a 0.2 meter interval, leading to 29 bins.
 For temporal information, we take all the 3D points from the past 5 timestamps. Thus our input  is a 4 dimensional tensor  consisting of time, height, X and Y.

For both our early-fusion and late-fusion models, we train from scratch using Adam optimizer \cite{kingma2014adam} with a learning rate of 1e-4. The model is trained on a 4 Titan XP GPU server with batch size of 12. We train the model for 100K iteration with learning rate halved at 60K and 80K iterations respectively.

\paragraph{Detection Results:} We compare  our model against state-of-the-art real-time detectors including SSD \cite{liu2016ssd}, MobileNet \cite{howard2017mobilenets} and SqueezeNet \cite{iandola2016squeezenet}. Note that these detectors   are all developed to work on 2D detection from  images. To make them competitive, we also build our predefined boxes into their system, which further easy the task for those detectors. The region of interest is $144\times 80M$ centered at ego-car during inference time. We keeps the same voxelization for all models and evaluate detections against ground truth vehicle bounding boxes with at minimum of three 3D points. Vehicles with less than three points are considered don't care regions.
We consider a detection correct if it has an IoU against any ground truth vehicle booundign box larger than 0.7.  Note that  for a vehicle with typical size of $3.5\times 6$ meters, 0.7 IoU means we can at most miss 0.35 meters along width and 0.6 meters along length. Fig.~\ref{fig:pr} shows the precision recall  curve for all approaches, where clearly our model is able to achieve higher recall, which is crucial for autonomous driving. Furthermore, Tab.~\ref{tab:pr} shows mAP using different IoU thresholds. We can see that our method is able to outperform all other methods. Particularly at IoU 0.7, we achieve 4.7\% higher mAP than MobileNet \cite{howard2017mobilenets} while being twice faster, and 5.2\% better than SSD \cite{liu2016ssd} with similar running time.

\begin{figure}[t]
\begin{center}
   \includegraphics[width=\linewidth]{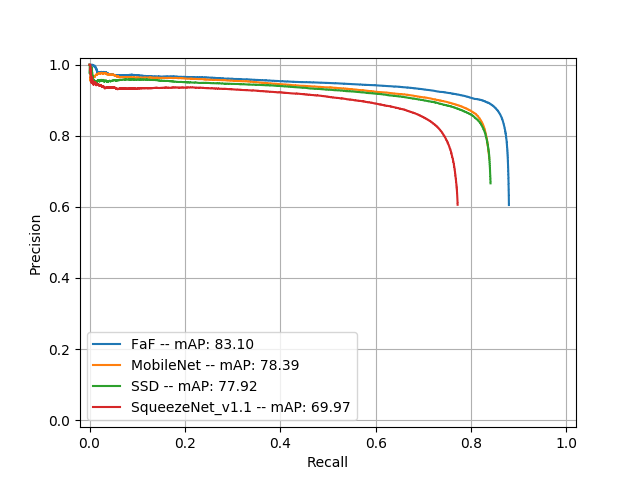}
\end{center}
   \caption{\textbf{P/R curve}}
\label{fig:pr}
\end{figure}

We also report performance as a function of the minimum number of 3D  points, which is used to filter ground truth bounding boxes during test time. Note that high level of sparsity is  due to occlusion or long distance vehicles. As shown in Fig.~\ref{fig:minpts}, our method is able to outperform other methods at all levels. We evaluate with a minimum of   0 point is, to show the importance of exploiting temporal information.

 We are also interested in knowing how the model perform as a function of  vehicle  distance. Towards this goal,  we extend the predictions to be as far as 100 meters away. Fig.~\ref{fig:range} shows the mAP with IoU 0.7 on vehicles within different distance ranges. We can see that all methods are doing well on nearby vehicles, while our method is significantly better  at  long range. Note that  all methods perform poorly at 100 meters  due to lack of 3D points at that distance.

 \begin{figure}[t]
 \begin{center}
    \includegraphics[width=\linewidth]{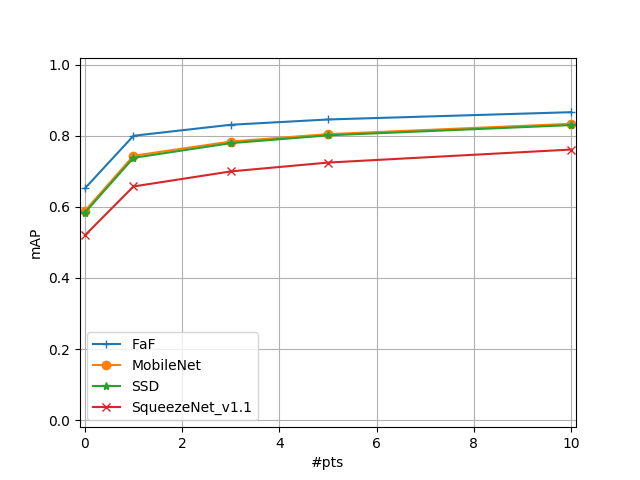}
 \end{center}
    \caption{\textbf{mAP on different number of minimum 3D points}}
 \label{fig:minpts}
 \end{figure}

 \begin{figure}[t]
 \begin{center}
    \includegraphics[width=\linewidth]{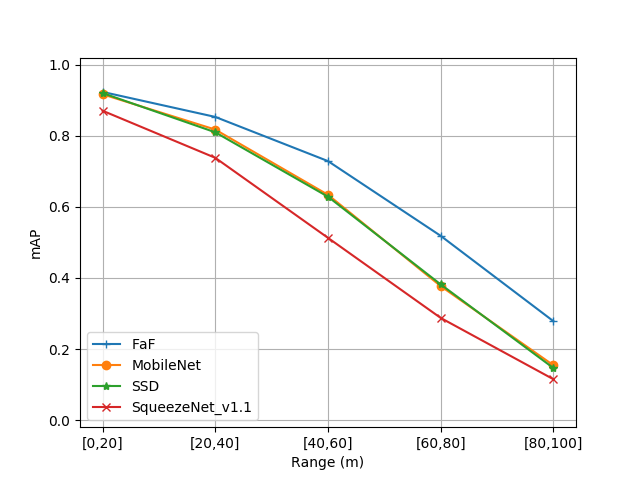}
 \end{center}
    \caption{\textbf{mAP over distance}}
 \label{fig:range}
 \end{figure}


\begin{table*}[t]
\setlength{\tabcolsep}{1mm}
\centering
\begin{tabular}{| c | c | c | c | c | c || c | c | c | c | c | c | c || c |}
		\hline
				{\scriptsize{Single}} & \scriptsize{5 Frames}  & \scriptsize{Early} & \scriptsize{Laster} & \scriptsize{w/ F} & \scriptsize{w/ T} & {IoU 0.5} & {IoU 0.6} & {IoU 0.7} & {IoU 0.8} & {IoU 0.9} & Time [ms] \\ \hline\hline
	\checkmark & & & & & &                                     89.81 & 86.27 & 77.20 & 52.28 & 6.33 & 9 \\ \hline
	& \checkmark & \checkmark & & & &                          91.49 & 88.57 & 80.90 & 57.14 & 8.17	& 11 \\ \hline
	& \checkmark & & \checkmark    & & &                       92.01 & 89.37 & 82.33 & 58.77 & 8.93 & 29	\\ \hline
	& \checkmark & & \checkmark & \checkmark & &               92.02 & 89.34 & 81.55 & 58.61 & 9.62	& 30	\\ \hline
 	& \checkmark & & \checkmark &\checkmark & \checkmark &     \bf93.24 & \bf90.54 & \bf83.10 & \bf61.61 & \bf11.83 & 30	\\ \hline
	\end{tabular}
	\caption{Ablation study, on $144\times 80$ region with vehicles having $\ge$3 number 3D points}
	\label{tab:abl}
\end{table*}

\paragraph{Ablation Study: }  We conducted ablation experiments within our framework to show how important each of the component is. We fixed the training setup for all experiments. As shown in Tab.~\ref{tab:abl}, using temporal information with early fusion gives 3.7\% improvement on mAP at IoU 0.7. While later fusion uses the same information as early fusion, it is able to get 1.4\% extra improvement as it can model more complex temporal features. In addition, adding prediction loss gives similar detection results on current frame alone, however it enables us to decode tracklets and provides evidence to output smoother detections, thus giving the best performance, \ie 6\% points better on mAP at IoU 0.7 than single frame detector.

\begin{figure}[t]
\begin{center}
   \includegraphics[width=\linewidth]{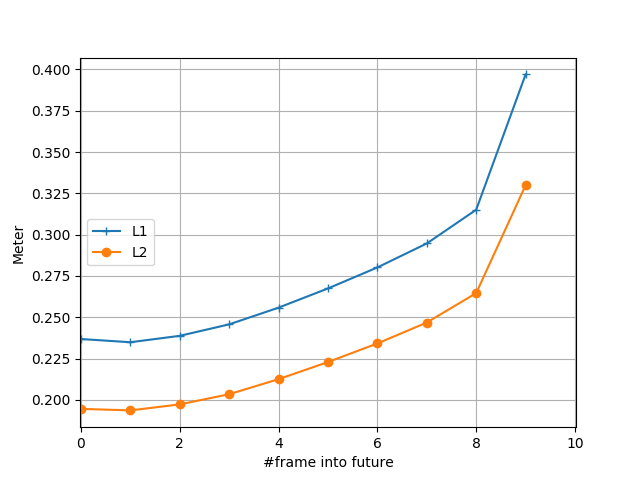}
\end{center}
   \caption{\textbf{Motion forecasting performance}}
\label{fig:pred}
\end{figure}

\begin{table}
\begin{center}
\begin{tabular}{|l|c|c|c|c|c|}
\hline
           & MOTA  & MOTP & MT & ML  \\ \hline\hline
FaF        & \bf 80.9     & 85.3 & \bf 75 & \bf 10.6 \\
Hungarian  & 73.1     & \bf 85.4 & 55.4  & 20.8 \\ \hline
\end{tabular}
\end{center}
\caption{Tracking performance}
\label{tab:tracking}
\end{table}

\begin{figure*}[t]
\small
\begin{center}
\setlength\tabcolsep{1.5pt}
\begin{tabular}{c c c c }
      \vspace{1.5pt}
      \begin{minipage}{.33\textwidth}
          \includegraphics[width=\textwidth]{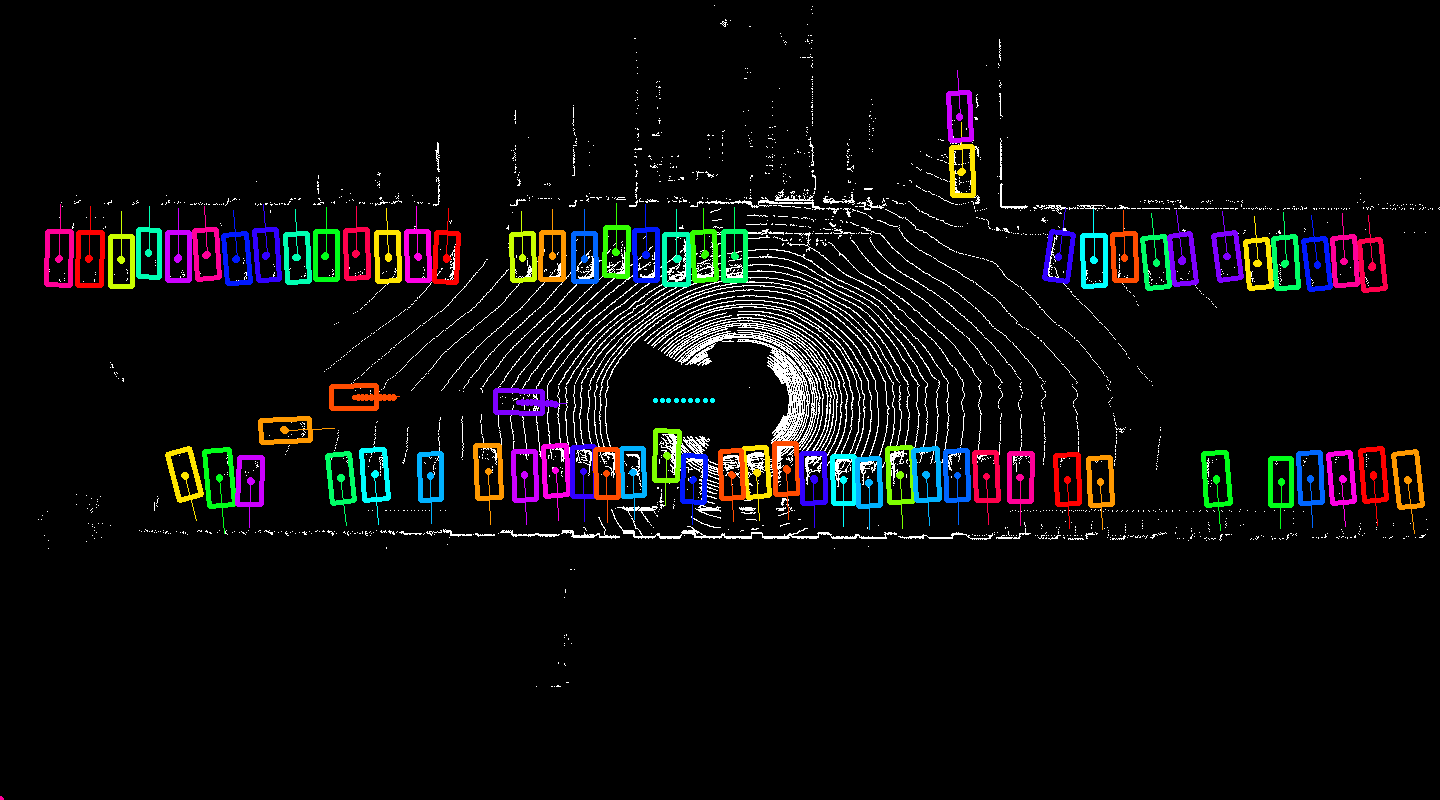}
      \end{minipage} &
      \begin{minipage}{.33\textwidth}
          \includegraphics[width=\linewidth]{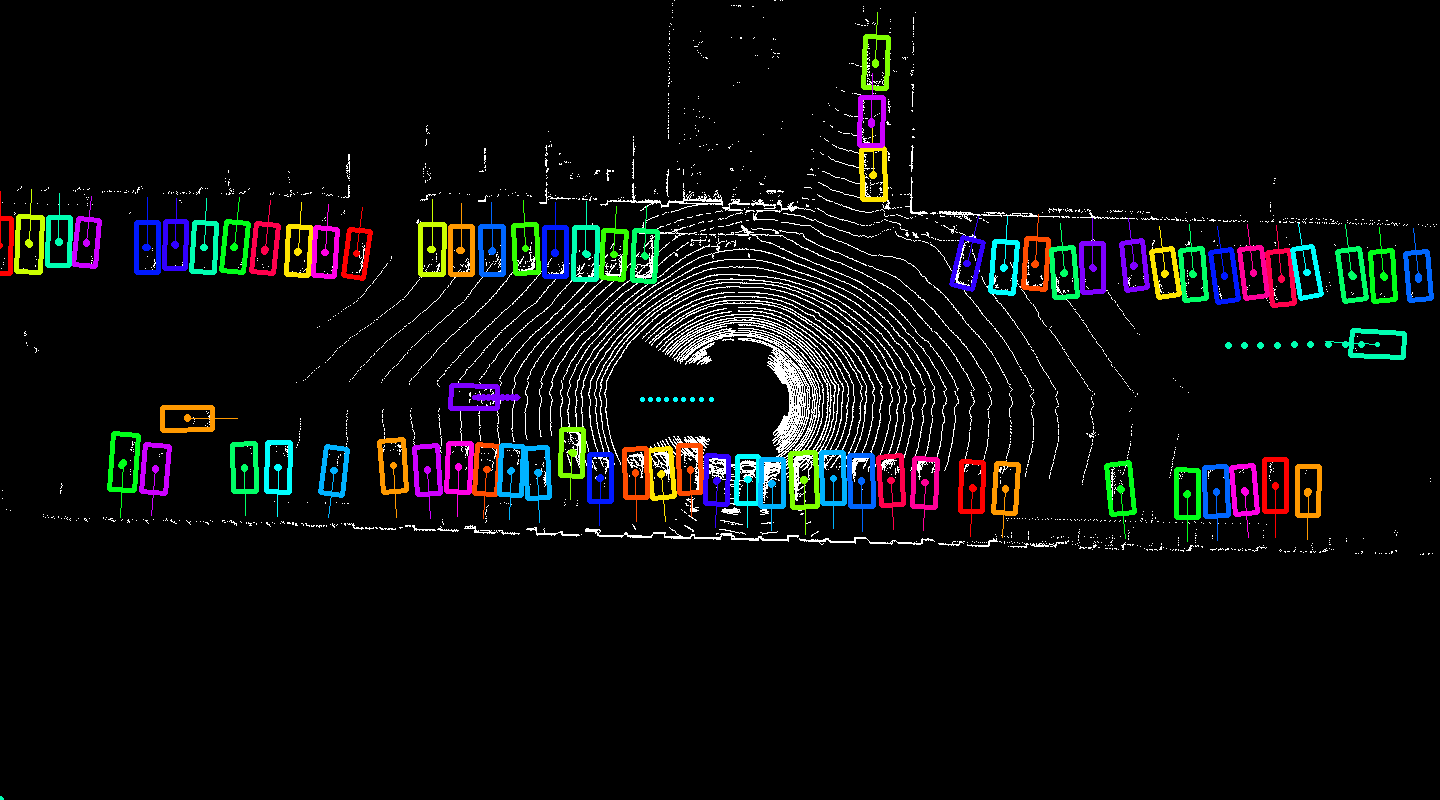}
      \end{minipage}&
      \begin{minipage}{.33\textwidth}
          \includegraphics[width=\linewidth]{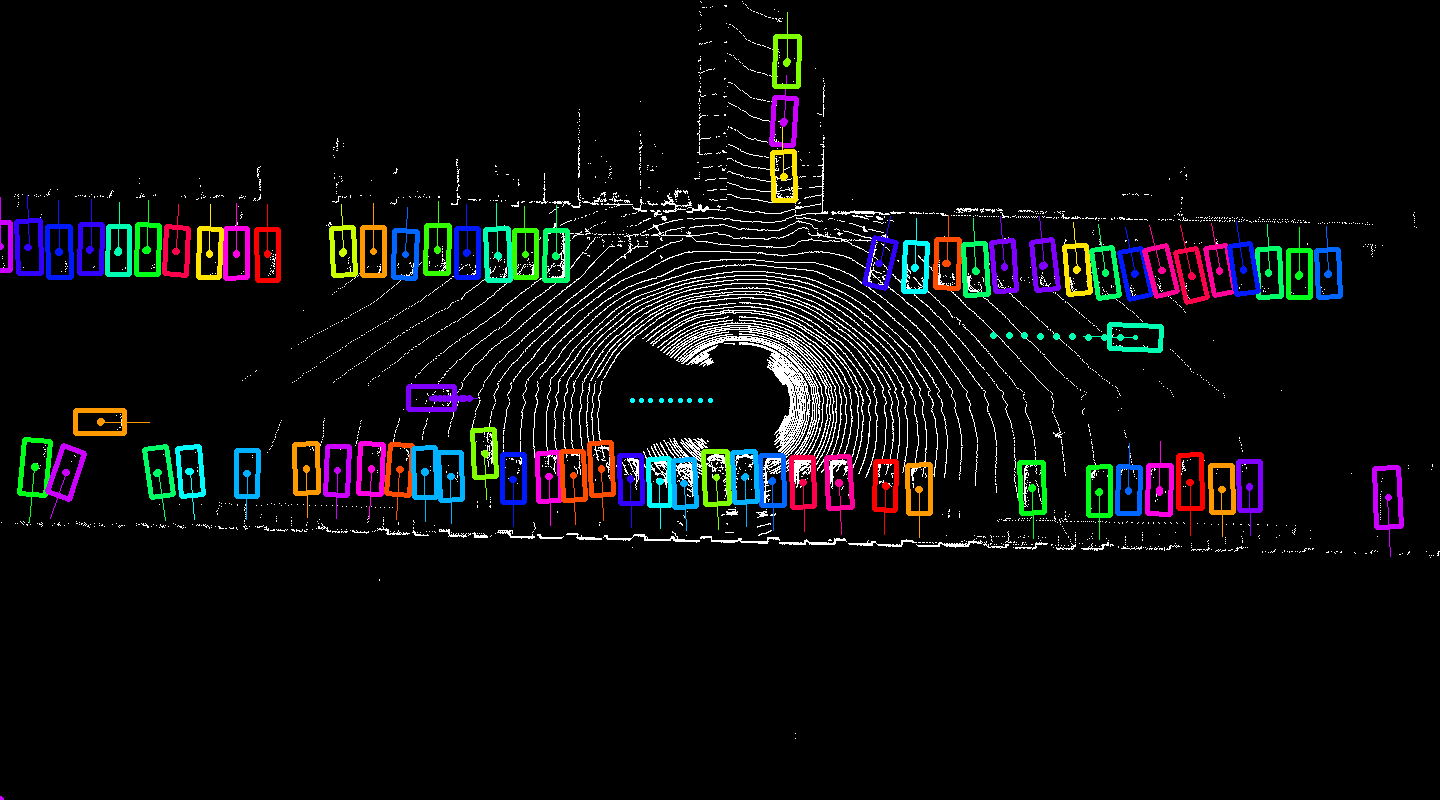}
      \end{minipage}\\

      \vspace{1.5pt}
      \begin{minipage}{.33\textwidth}
          \includegraphics[width=\textwidth]{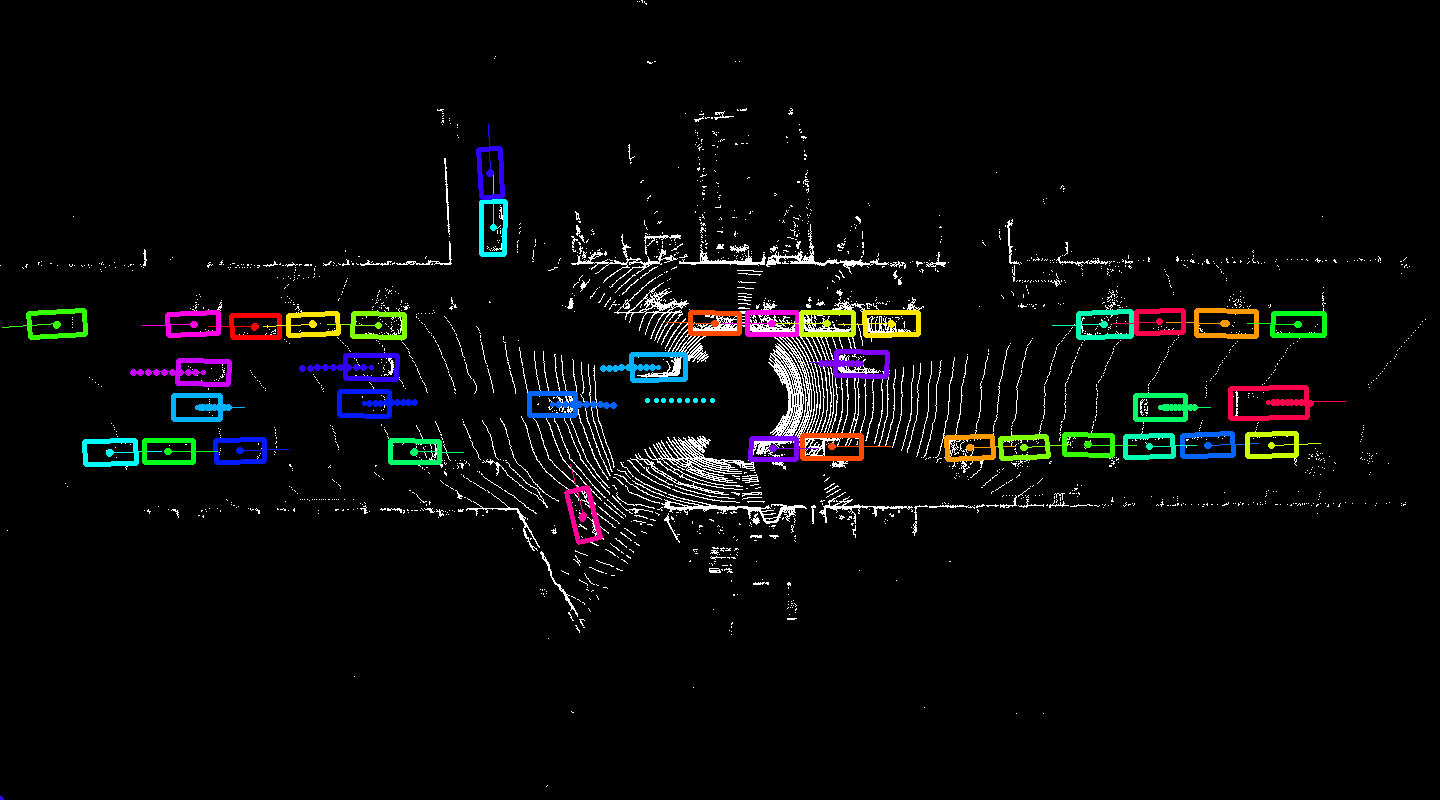}
      \end{minipage} &
      \begin{minipage}{.33\textwidth}
          \includegraphics[width=\linewidth]{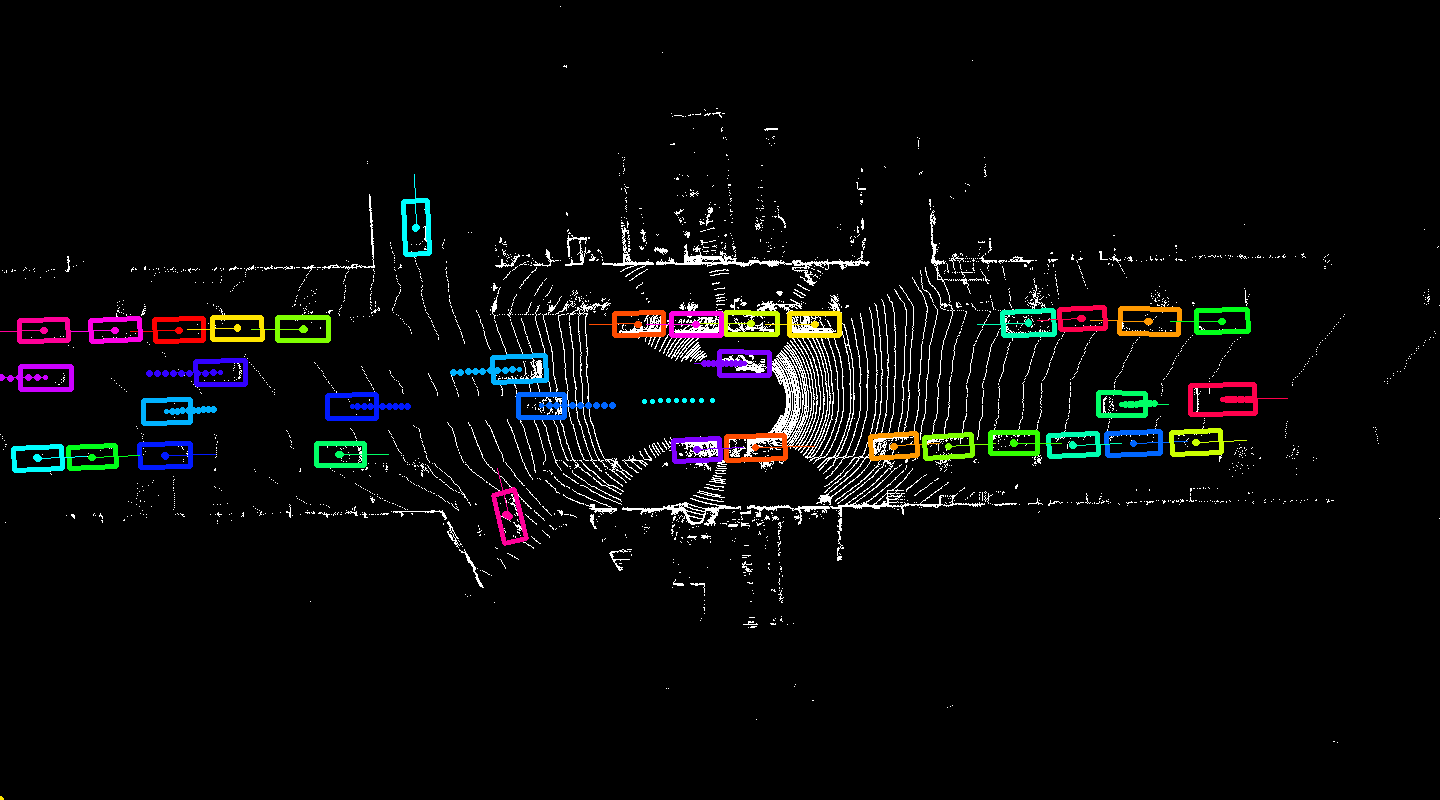}
      \end{minipage}&
      \begin{minipage}{.33\textwidth}
          \includegraphics[width=\linewidth]{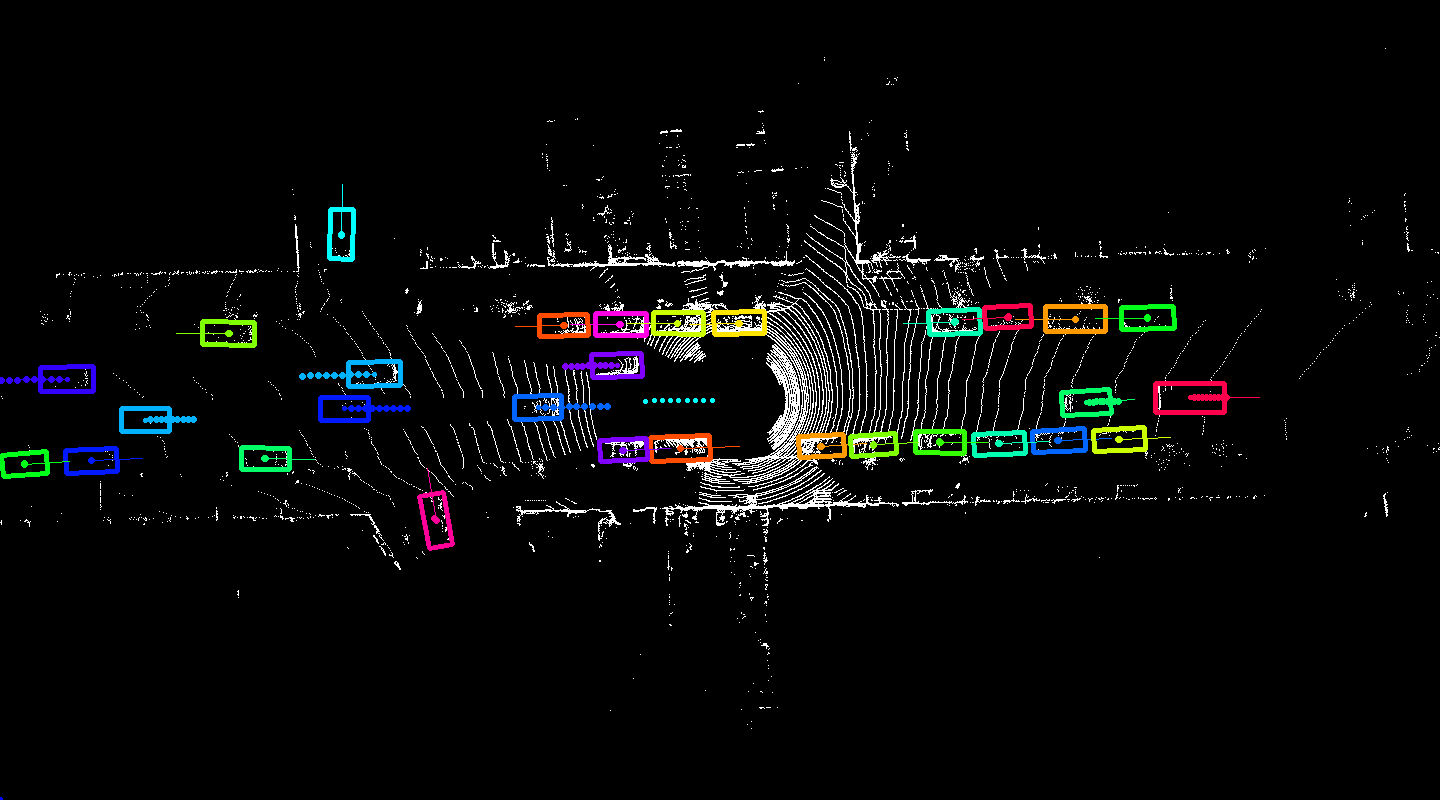}
      \end{minipage}\\

      \vspace{1.5pt}
      \begin{minipage}{.33\textwidth}
          \includegraphics[width=\textwidth]{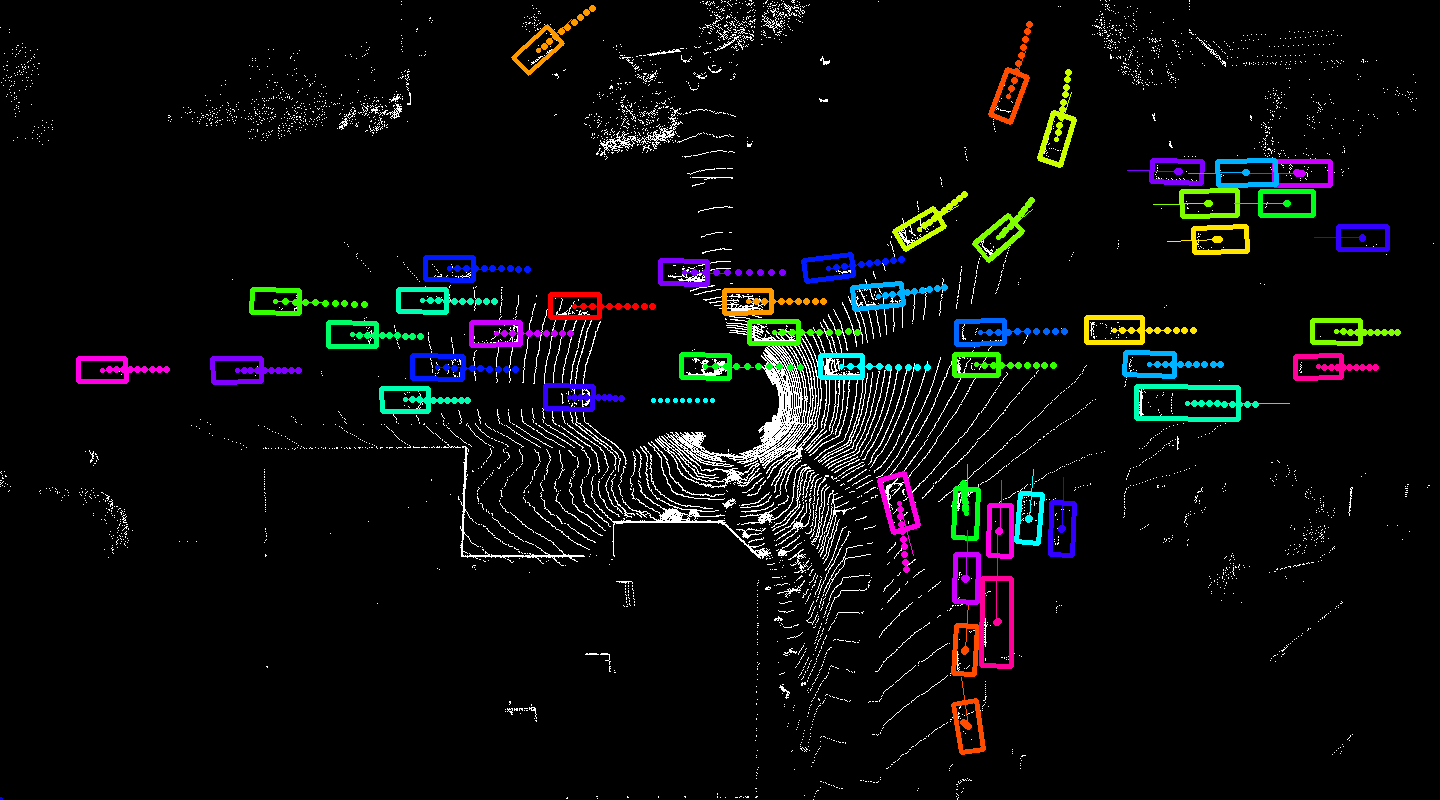}
      \end{minipage} &
      \begin{minipage}{.33\textwidth}
        \includegraphics[width=\textwidth]{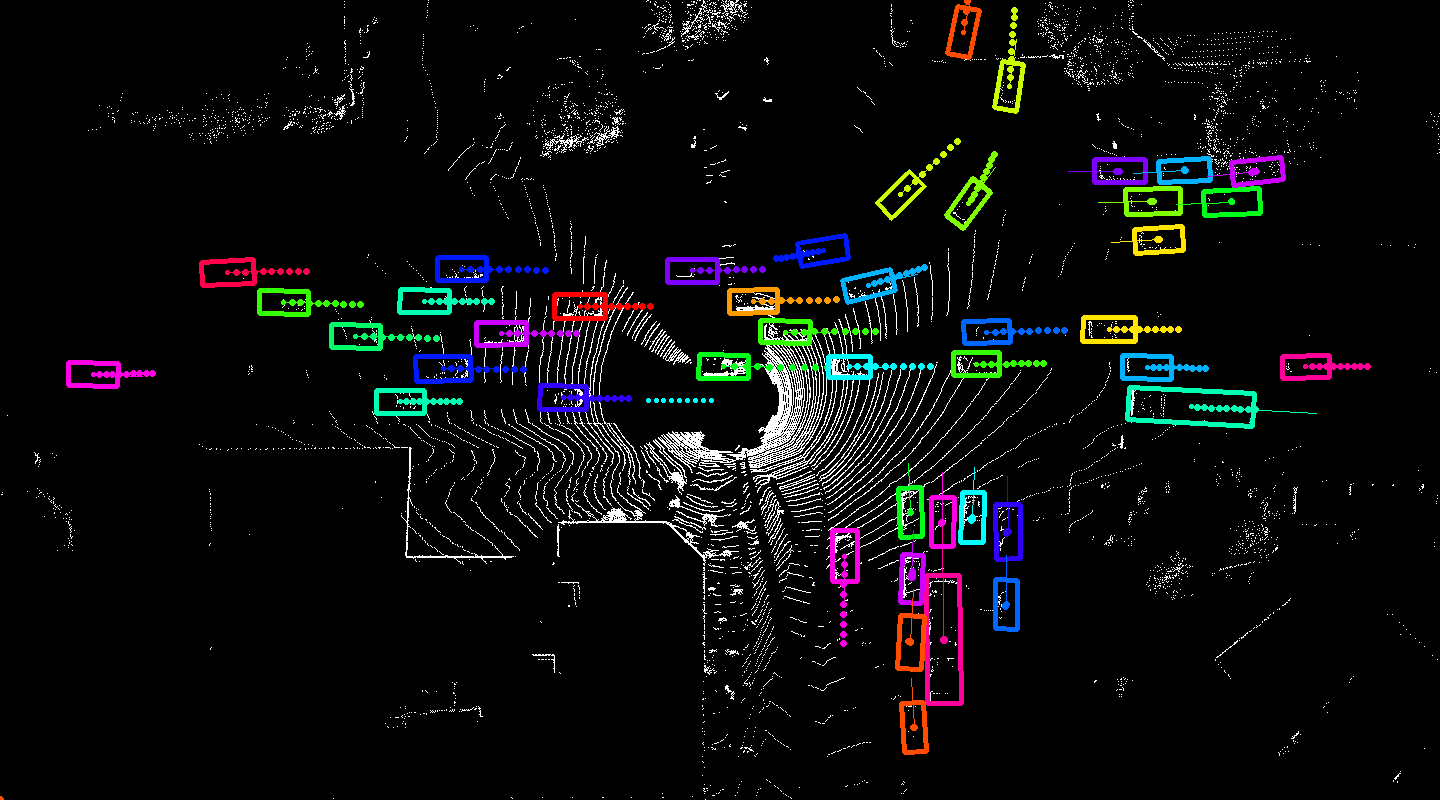}
      \end{minipage}&
      \begin{minipage}{.33\textwidth}
        \includegraphics[width=\textwidth]{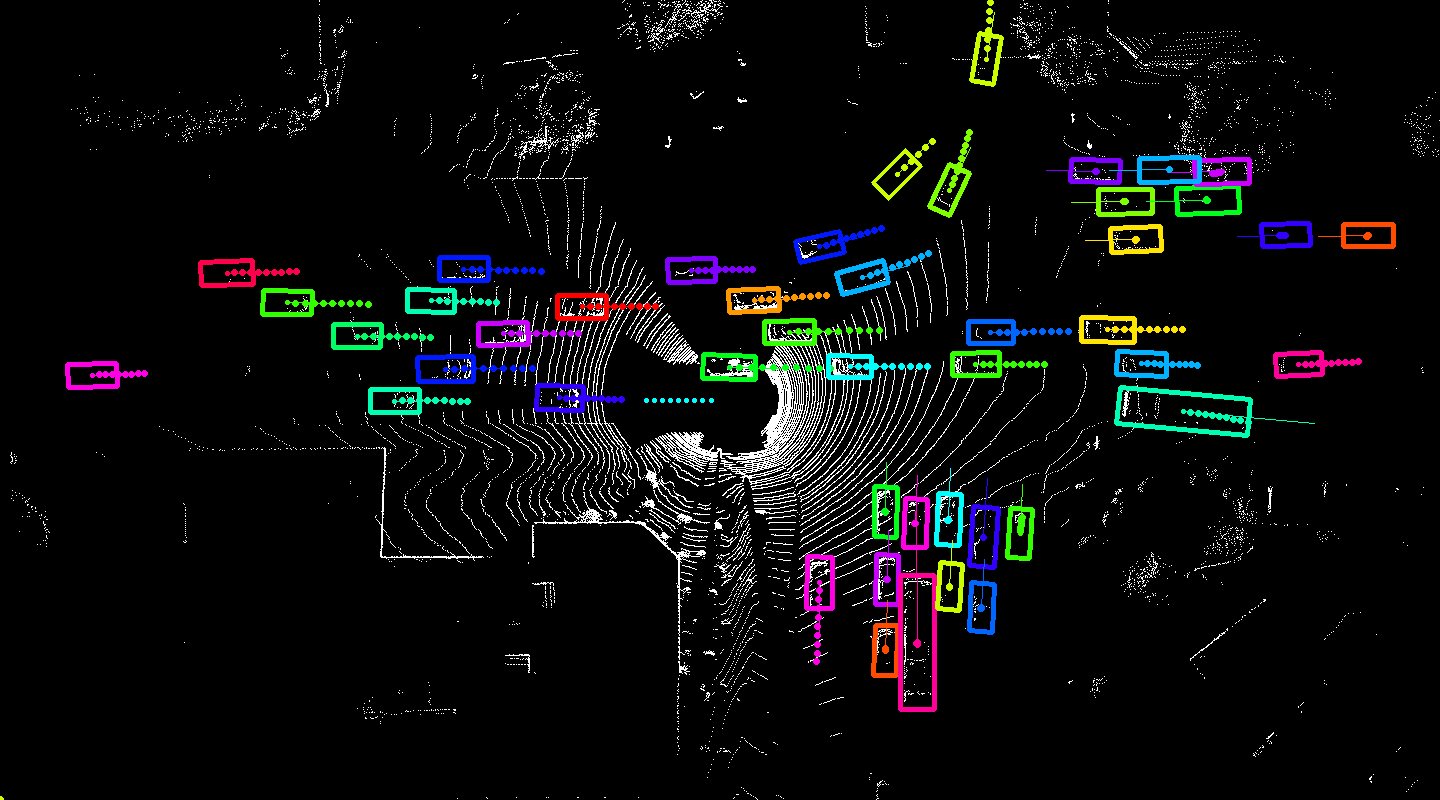}
      \end{minipage}\\

      \vspace{1.5pt}
      \begin{minipage}{.33\textwidth}
          \includegraphics[width=\textwidth]{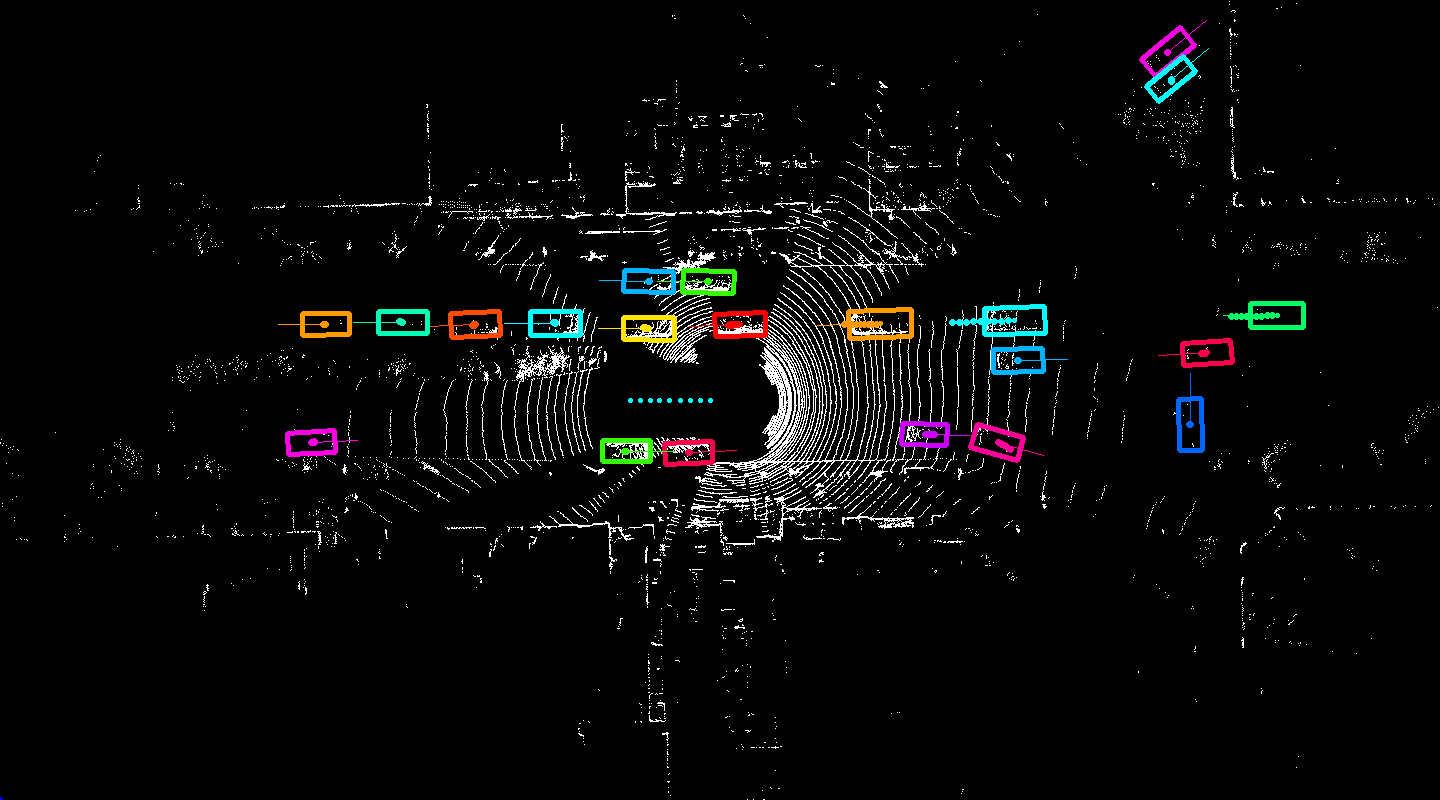}
      \end{minipage} &
      \begin{minipage}{.33\textwidth}
        \includegraphics[width=\textwidth]{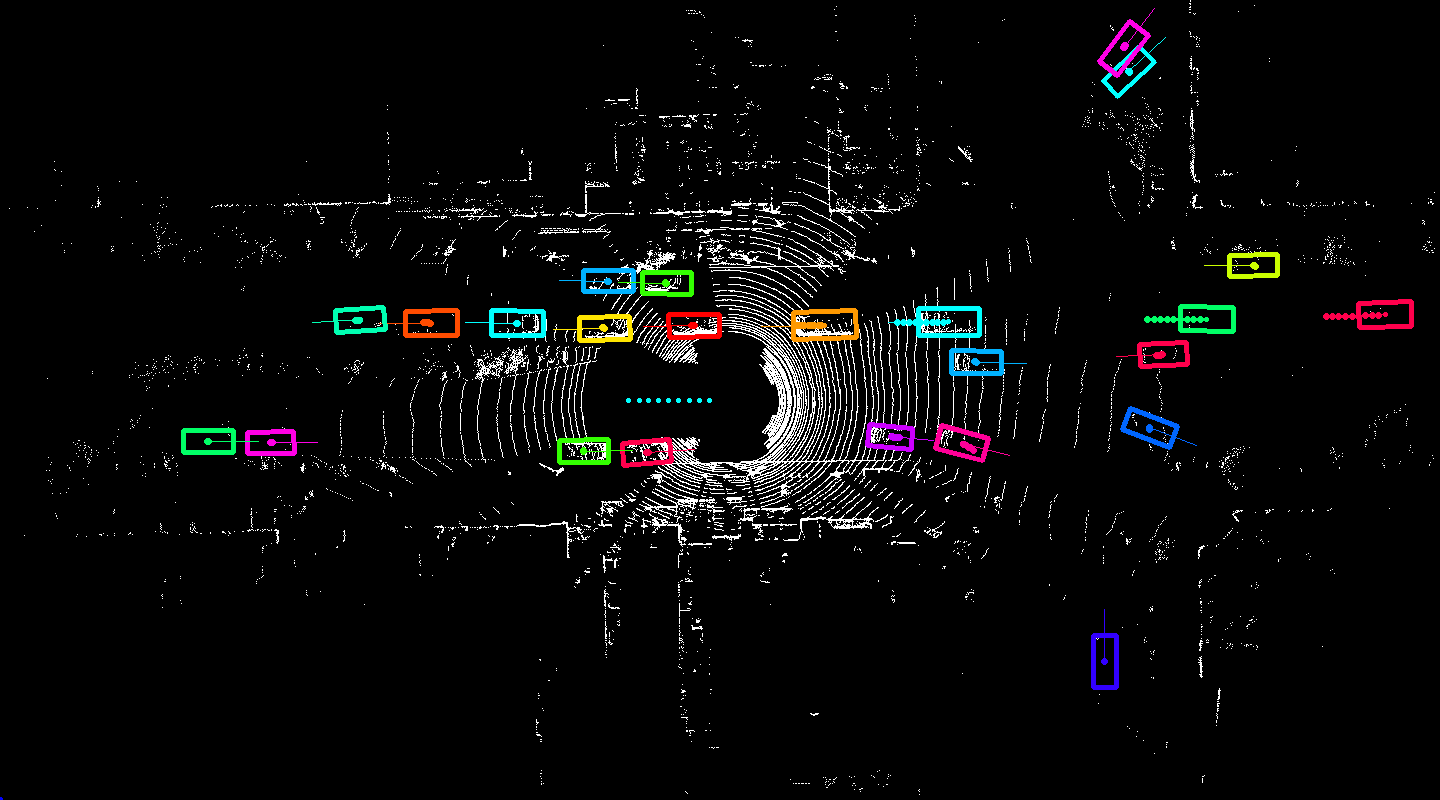}
      \end{minipage}&
      \begin{minipage}{.33\textwidth}
        \includegraphics[width=\textwidth]{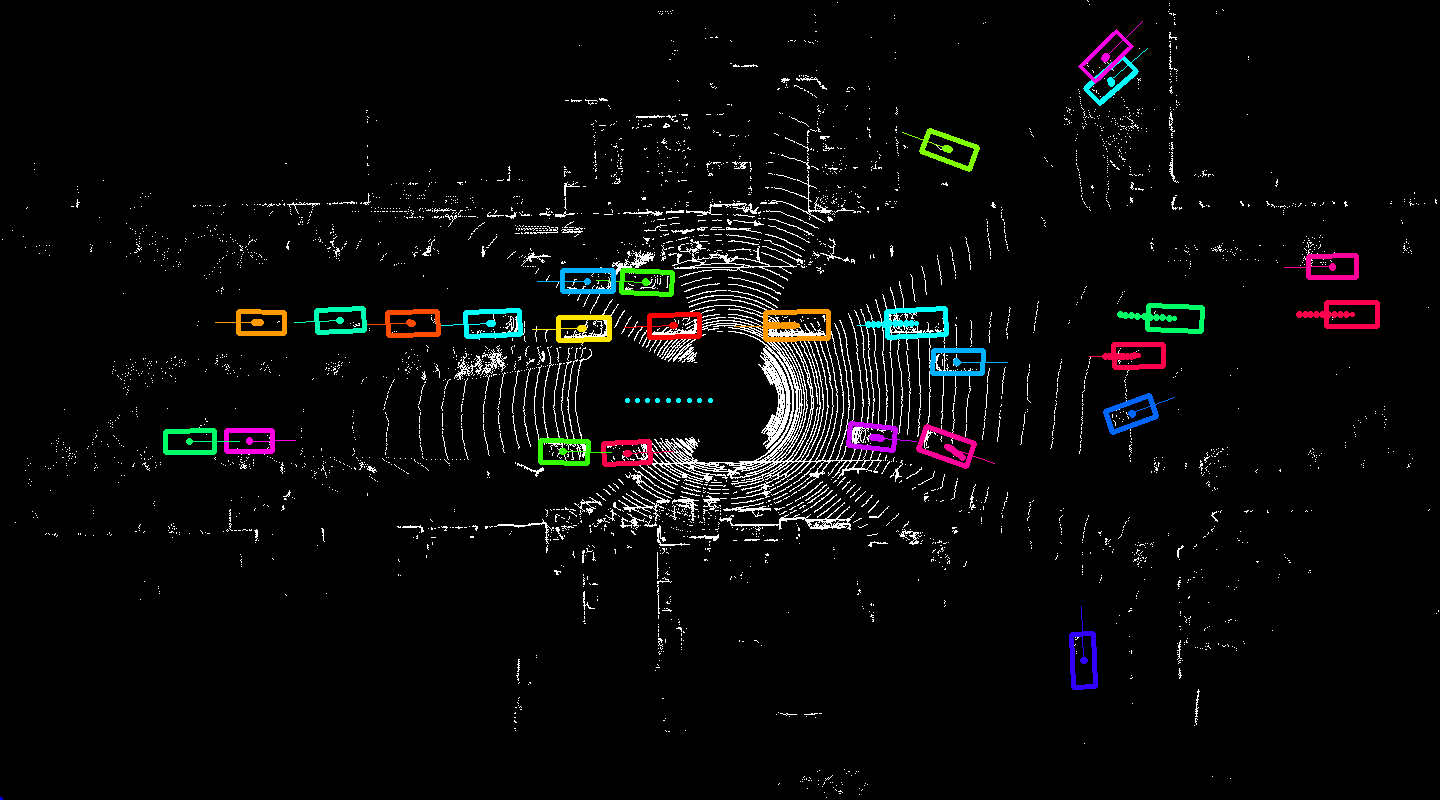}
      \end{minipage}\\

\end{tabular}
\end{center}
   \caption{\textbf{Qualitative results on 144x80M region [best view in color].} Same color represents same vehicle across different timeframe. Each vehicle has `dot' presents the center locations of current and future time frames}
 \label{fig:he_vis}
\end{figure*}

\paragraph{Tracking:} Our model is able to output detections with track ids directly. We evaluate the raw tracking output without adding any sophisticated tracking pipeline on top. Tab.~\ref{tab:tracking} shows the comparison between our model's output and a Hungarian method on top of our detection results. We follow the KITTI protocol  \cite{Geiger2012CVPR} and compute  MOTA, MOTP, Mostly-Tracked (MT) and  Mostly-Lost (ML) across all 100 validation sequences. The evaluation script uses IoU 0.5 for association and 0.9 score for thresholding both methods. We can see that our final output achieves $80.9\%$ in MOTA, $7.8\%$ better than Hungarian, as well as $20\%$ better on MT, $10\%$ lower on ML, while still having similar MOTP.

\paragraph{Motion Forecasting:} We evaluate the forecasting ability of the model by computing the average L1 and L2 distances of the vehicles' center location. As shown in Fig.~\ref{fig:pred}, we are able to predict  10 frames into the future with L2 distance only less than 0.33 meter.
Note that  due to the nature of the problem, we can only evaluate on true positives, which in our case has  a corresponding recall of $92.5\%$.

\paragraph{Qualitative Results:}  Fig.~\ref{fig:he_vis}  shows our results on a $144\times 80$ meters region. We provide 4 sequences, where the top 3 rows show that our model is able to perform well at complex scenes, giving accurate rotated bounding boxes on both small vehicles as well big trucks. Note that  our model also gives accurate motion forecasting for both fast moving vehicles and static vehicles (where all future center locations overlay at the current location).
The last row shows one failure case, where our detector fails on the center right blue vehicle. This is due to  the sparsity of the  3D points.

\section{Conclusion}

We have proposed a holistic model that reasons jointly about  detection, prediction and tracking in the scenario of autonomous driving. We show that it runs real-time and achieves very good accuracy on all tasks. 
In the future, we plan to incorporate RoI align  in order to have better feature representations. We also plan to test other categories such as pedestrians and produce longer term predictions.

{\small
\bibliographystyle{ieee}
\bibliography{egbib}

\begin{thebibliography}{10}\itemsep=-1pt

\bibitem{alahi2016social}
A.~Alahi, K.~Goel, V.~Ramanathan, A.~Robicquet, L.~Fei-Fei, and S.~Savarese.
\newblock Social lstm: Human trajectory prediction in crowded spaces.
\newblock In {\em Proceedings of the IEEE Conference on Computer Vision and
  Pattern Recognition}, pages 961--971, 2016.

\bibitem{chen20173d}
X.~Chen, K.~Kundu, Y.~Zhu, H.~Ma, S.~Fidler, and R.~Urtasun.
\newblock 3d object proposals using stereo imagery for accurate object class
  detection.
\newblock {\em IEEE Transactions on Pattern Analysis and Machine Intelligence},
  2017.

\bibitem{chen2016multi}
X.~Chen, H.~Ma, J.~Wan, B.~Li, and T.~Xia.
\newblock Multi-view 3d object detection network for autonomous driving.
\newblock {\em arXiv preprint arXiv:1611.07759}, 2016.

\bibitem{dai2016r}
J.~Dai, Y.~Li, K.~He, and J.~Sun.
\newblock R-fcn: Object detection via region-based fully convolutional
  networks.
\newblock In {\em Advances in neural information processing systems}, pages
  379--387, 2016.

\bibitem{feichtenhofer2017detect}
C.~Feichtenhofer, A.~Pinz, and A.~Zisserman.
\newblock Detect to track and track to detect.
\newblock {\em arXiv preprint arXiv:1710.03958}, 2017.

\bibitem{Geiger2012CVPR}
A.~Geiger, P.~Lenz, and R.~Urtasun.
\newblock Are we ready for autonomous driving? the kitti vision benchmark
  suite.
\newblock In {\em Conference on Computer Vision and Pattern Recognition
  (CVPR)}, 2012.

\bibitem{gong2011multi}
H.~Gong, J.~Sim, M.~Likhachev, and J.~Shi.
\newblock Multi-hypothesis motion planning for visual object tracking.
\newblock In {\em Computer Vision (ICCV), 2011 IEEE International Conference
  on}, pages 619--626. IEEE, 2011.

\bibitem{he2017maskrcnn}
K.~He, G.~Gkioxari, P.~Doll\'{a}r, and R.~Girshick.
\newblock {Mask R-CNN}.
\newblock {\em arXiv preprint arXiv:1703.06870}, 2017.

\bibitem{held2016learning}
D.~Held, S.~Thrun, and S.~Savarese.
\newblock Learning to track at 100 fps with deep regression networks.
\newblock In {\em European Conference on Computer Vision}, pages 749--765.
  Springer, 2016.

\bibitem{howard2017mobilenets}
A.~G. Howard, M.~Zhu, B.~Chen, D.~Kalenichenko, W.~Wang, T.~Weyand,
  M.~Andreetto, and H.~Adam.
\newblock Mobilenets: Efficient convolutional neural networks for mobile vision
  applications.
\newblock {\em arXiv preprint arXiv:1704.04861}, 2017.

\bibitem{huang2016speed}
J.~Huang, V.~Rathod, C.~Sun, M.~Zhu, A.~Korattikara, A.~Fathi, I.~Fischer,
  Z.~Wojna, Y.~Song, S.~Guadarrama, et~al.
\newblock Speed/accuracy trade-offs for modern convolutional object detectors.
\newblock {\em arXiv preprint arXiv:1611.10012}, 2016.

\bibitem{iandola2016squeezenet}
F.~N. Iandola, S.~Han, M.~W. Moskewicz, K.~Ashraf, W.~J. Dally, and K.~Keutzer.
\newblock Squeezenet: Alexnet-level accuracy with 50x fewer parameters and< 0.5
  mb model size.
\newblock {\em arXiv preprint arXiv:1602.07360}, 2016.

\bibitem{kingma2014adam}
D.~Kingma and J.~Ba.
\newblock Adam: A method for stochastic optimization.
\newblock {\em arXiv preprint arXiv:1412.6980}, 2014.

\bibitem{lee2017desire}
N.~Lee, W.~Choi, P.~Vernaza, C.~B. Choy, P.~H. Torr, and M.~Chandraker.
\newblock Desire: Distant future prediction in dynamic scenes with interacting
  agents.
\newblock {\em arXiv preprint arXiv:1704.04394}, 2017.

\bibitem{li20163d}
B.~Li.
\newblock 3d fully convolutional network for vehicle detection in point cloud.
\newblock {\em arXiv preprint arXiv:1611.08069}, 2016.

\bibitem{lin2017focal}
T.-Y. Lin, P.~Goyal, R.~Girshick, K.~He, and P.~Doll{\'a}r.
\newblock Focal loss for dense object detection.
\newblock {\em arXiv preprint arXiv:1708.02002}, 2017.

\bibitem{liu2016ssd}
W.~Liu, D.~Anguelov, D.~Erhan, C.~Szegedy, S.~Reed, C.-Y. Fu, and A.~C. Berg.
\newblock {SSD}: Single shot multibox detector.
\newblock In {\em ECCV}, 2016.

\bibitem{ma2015hierarchical}
C.~Ma, J.-B. Huang, X.~Yang, and M.-H. Yang.
\newblock Hierarchical convolutional features for visual tracking.
\newblock In {\em Proceedings of the IEEE International Conference on Computer
  Vision}, pages 3074--3082, 2015.

\bibitem{ma2017forecasting}
W.-C. Ma, D.-A. Huang, N.~Lee, and K.~M. Kitani.
\newblock Forecasting interactive dynamics of pedestrians with fictitious play.
\newblock In {\em Proceedings of the IEEE Conference on Computer Vision and
  Pattern Recognition}, pages 774--782, 2017.

\bibitem{mathieu2015deep}
M.~Mathieu, C.~Couprie, and Y.~LeCun.
\newblock Deep multi-scale video prediction beyond mean square error.
\newblock {\em arXiv preprint arXiv:1511.05440}, 2015.

\bibitem{nam2016learning}
H.~Nam and B.~Han.
\newblock Learning multi-domain convolutional neural networks for visual
  tracking.
\newblock In {\em Proceedings of the IEEE Conference on Computer Vision and
  Pattern Recognition}, pages 4293--4302, 2016.

\bibitem{pellegrini2009you}
S.~Pellegrini, A.~Ess, K.~Schindler, and L.~Van~Gool.
\newblock You'll never walk alone: Modeling social behavior for multi-target
  tracking.
\newblock In {\em Computer Vision, 2009 IEEE 12th International Conference on},
  pages 261--268. IEEE, 2009.

\bibitem{redmon2016yolo9000}
J.~Redmon and A.~Farhadi.
\newblock Yolo9000: better, faster, stronger.
\newblock {\em arXiv preprint arXiv:1612.08242}, 2016.

\bibitem{ren2015faster}
S.~Ren, K.~He, R.~Girshick, and J.~Sun.
\newblock Faster {R-CNN}: Towards real-time object detection with region
  proposal networks.
\newblock In {\em Neural Information Processing Systems ({NIPS})}, 2015.

\bibitem{simonyan2014very}
K.~Simonyan and A.~Zisserman.
\newblock Very deep convolutional networks for large-scale image recognition.
\newblock {\em arXiv preprint arXiv:1409.1556}, 2014.

\bibitem{srivastava2015unsupervised}
N.~Srivastava, E.~Mansimov, and R.~Salakhudinov.
\newblock Unsupervised learning of video representations using lstms.
\newblock In {\em International Conference on Machine Learning}, pages
  843--852, 2015.

\bibitem{tao2016siamese}
R.~Tao, E.~Gavves, and A.~W. Smeulders.
\newblock Siamese instance search for tracking.
\newblock In {\em Proceedings of the IEEE Conference on Computer Vision and
  Pattern Recognition}, pages 1420--1429, 2016.

\bibitem{walker2016uncertain}
J.~Walker, C.~Doersch, A.~Gupta, and M.~Hebert.
\newblock An uncertain future: Forecasting from static images using variational
  autoencoders.
\newblock In {\em European Conference on Computer Vision}, pages 835--851.
  Springer, 2016.

\bibitem{wang2015visual}
L.~Wang, W.~Ouyang, X.~Wang, and H.~Lu.
\newblock Visual tracking with fully convolutional networks.
\newblock In {\em Proceedings of the IEEE International Conference on Computer
  Vision}, pages 3119--3127, 2015.

\bibitem{wang2013learning}
N.~Wang and D.-Y. Yeung.
\newblock Learning a deep compact image representation for visual tracking.
\newblock In {\em Advances in neural information processing systems}, pages
  809--817, 2013.

\bibitem{wu2016squeezedet}
B.~Wu, F.~Iandola, P.~H. Jin, and K.~Keutzer.
\newblock Squeezedet: Unified, small, low power fully convolutional neural
  networks for real-time object detection for autonomous driving.
\newblock {\em arXiv preprint arXiv:1612.01051}, 2016.

\end{thebibliography}
}

\end{document}